%% file: main.tex
\documentclass{article}

\PassOptionsToPackage{square,numbers,comma}{natbib}

\usepackage[preprint]{neurips_2024}

\usepackage[utf8]{inputenc} %
\usepackage[T1]{fontenc}    %
\usepackage{hyperref}       %
\usepackage{url}            %
\usepackage{booktabs}       %
\usepackage{amsfonts}       %
\usepackage{nicefrac}       %
\usepackage{microtype}      %
\usepackage{xcolor}         %
\usepackage{graphicx}
\usepackage{amsmath}
\usepackage{caption}
\usepackage[shortlabels]{enumitem}
\usepackage[super]{nth}

\definecolor{midnightblue}{rgb}{0.1, 0.1, 0.44}

\usepackage{subcaption}
\title{Survival of the Fittest Representation: A Case Study with Modular Addition}

\newcommand{\mat}[1]{\mathbf{#1}}

\author{%
Xiaoman Delores Ding$^*$,\, Zifan Carl Guo$^*$, \,Eric J. Michaud, \,Ziming Liu$^\dagger$, \,Max Tegmark$^\dagger$ \\
Massachusetts Institute of Technology \\
\texttt{\{delores,carlguo,ericjm,zmliu,tegmark\}@mit.edu} \\
}

\begin{document}

\maketitle
\def\thefootnote{*}\footnotetext{Equal contribution. Order determined by a coin flip.}\def\thefootnote{\arabic{footnote}}\def\thefootnote{$\dagger$}\footnotetext{Equal advising.}\def\thefootnote{\arabic{footnote}}

\begin{abstract}\label{sec:abstract}

When a neural network can learn multiple distinct algorithms to solve a task, how does it ``choose'' between them during training? 
To approach this question, we take inspiration from ecology: when multiple species coexist, they eventually reach an equilibrium where some survive while others die out. 
Analogously, we suggest that a neural network at initialization contains many solutions (representations and algorithms), which compete with each other under pressure from resource constraints, with the ``fittest'' ultimately prevailing. To investigate this \emph{Survival of the Fittest hypothesis}, we conduct a case study on neural networks performing modular addition, and find that these networks' multiple circular representations at different Fourier frequencies undergo such competitive dynamics, with only a few circles surviving at the end. We find that the frequencies with high initial signals and gradients, the ``fittest,'' are more likely to survive. By increasing the embedding dimension, we also observe more surviving frequencies. Inspired by the Lotka-Volterra equations describing the dynamics between species, we find that the dynamics of the circles can be nicely characterized by a set of linear differential equations. Our results with modular addition show that it is possible to decompose complicated representations into simpler components, along with their basic interactions, to offer insight on the training dynamics of representations.~\footnote{Our code is available at \href{https://github.com/carlguo866/circle-survival}{https://github.com/carlguo866/circle-survival}.}

\end{abstract}

\section{Introduction}\label{sec:introduction}

The field of \emph{mechanistic interpretability} attempts to reverse-engineer the algorithms that neural networks learn. This involves understanding the representations (features) networks learn~\cite{liu2022towards,zou2023representation, cunningham2023sparse, bricken2023monosemanticity, gurnee2024language, marks2023geometry}
and how these play a role in larger circuits~\cite{olsson2022context, nanda2023progress, olah2020zoom, elhage2021mathematical, marks2024sparse}. While most such work studies networks as static objects, some have recently begun to study how network representations and circuits form over training~\cite{liu2022towards, olsson2022context, nanda2023progress, hoogland2024developmental, chen2023sudden, singh2024needs}. In studying training dynamics, we hope to understand not just \emph{what} neural networks learn, but \emph{how} and ultimately \emph{why} they learn the algorithms that they learn. This understanding may eventually be useful for training models with the properties we desire, such as improved efficiency and safety.

One broad question in mechanistic interpretability regards \emph{universality} \citep{olah2020zoom}: can models consistently learn the same algorithms across different seeds and scales? While some work has found evidence of universality \citep{olsson2022context,gould2024successor,gurnee2024universal}, in other cases there seems to be some variability in algorithms and representations networks learn to solve particular tasks~\cite{zhong2023the,mccoy2019berts,lampinen2024learned}. In this work, we ask: when networks have a choice between different representations, how do they choose which one to learn?

\begin{figure}
    \centering

    \begin{subfigure}[t]{\textwidth}
        \includegraphics[width=\linewidth]{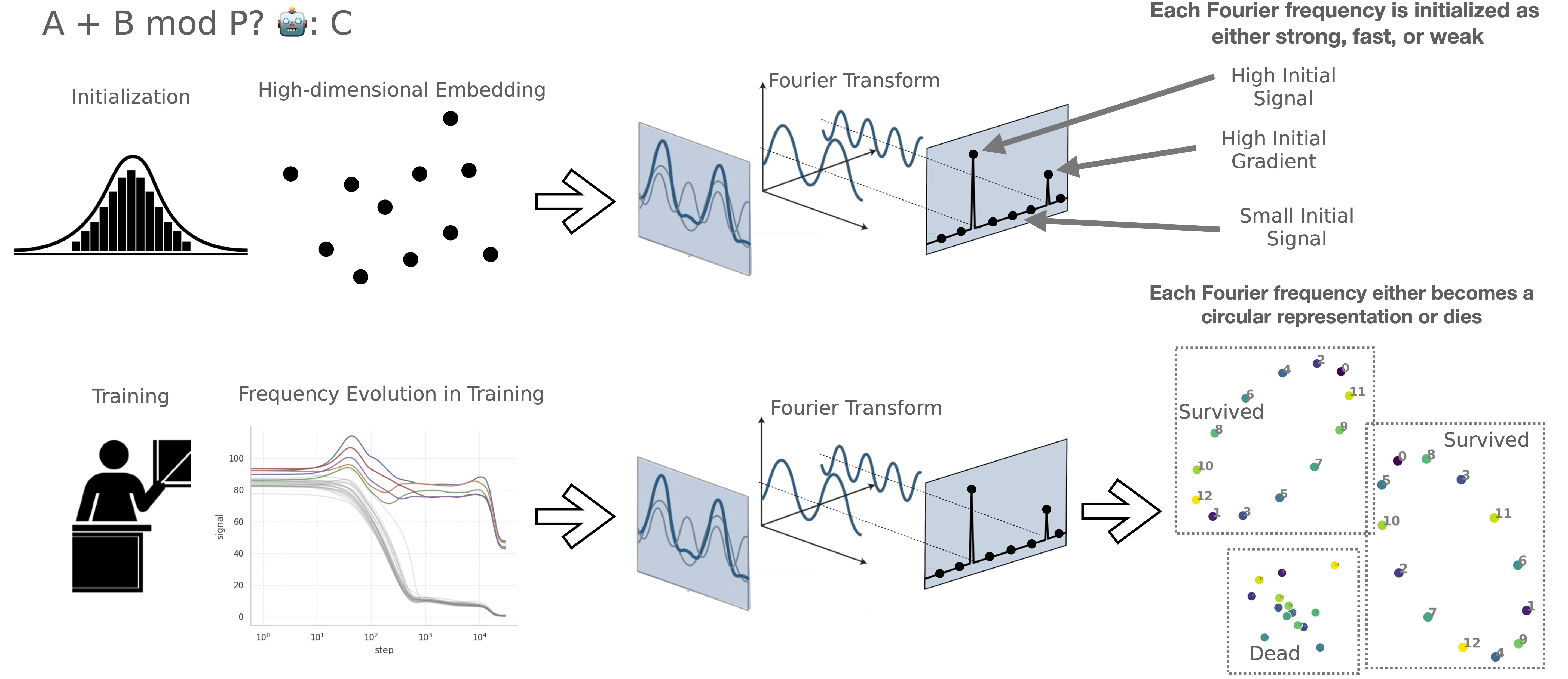}
    \end{subfigure}
    \caption{Our experiment method: perform a Fourier transform to the embedding, analyze the initialization, signal evolution, and their effects on the final learned circles over different training runs.}
    \label{fig:figure1}
\end{figure}

The question above is hard to answer since representations in the general case are high-dimensional and difficult to disentangle, let alone to understand the dynamics between multiple representations. Therefore, our study focuses on the toy models performing modular addition $a+b=c\ ({\rm mod}\ p)$, a mathematically defined and well studied problem. Prior works have shown that circular representations in the embedding, akin to how numbers are arranged around a clock, are key for modular addition models to generalize~\citep{liu2022towards,liu2023omnigrok,nanda2023progress,zhong2023the,power2022grokking}, and that the model learns a few such circles that nicely correspond to different Fourier frequencies~\citep{nanda2023progress}. Since the embedding is randomly initialized, the projection of the embedding onto the Fourier basis results in roughly similar frequencies, meaning that all possible frequencies have a chance to become the final circle representation. However, only 3-5 circles survive after training, while some frequencies do not form circles, as illustrated in Figure \ref{fig:figure1}. This begs the question: is there any pattern to how these representations form? 

To understand the pattern, we propose the \emph{Survival of the Fittest hypothesis} in analogy to ecosystems, where circles of different frequencies can be thought of as "species" competing for a fixed amount of total resources. The two systems are analogous in many aspects: (1) We find surviving circles to have large magnitudes and/or large expanding velocity (gradient) at initialization, supporting the ``survival of the fittest'' hypothesis. (2) The number of surviving frequencies increases as the resources (embedding dimension) increases. (3) A linear differential equation, inspired by the Lotka–Volterra equations \citep{alon2019introduction} describing interacting dynamics between species, is able to fit the evolution of the magnitude of these circles quite well, even when we push the interaction matrices to be extremely sparse via Lasso regression. 
Some circle pairs display collaborative behavior, while other pairs are competitive.

Our results show that it is possible to decompose complicated, high-dimensional embeddings into low-dimensional, interpretable representations that also have simple interactions. Our work serves as a proof-of-concept example, demonstrating that this decomposition-style analysis can help understand model training dynamics in more real-world contexts. The paper is organized as follows: In Section~\ref{sec:problem-setup}, we introduce the problem setup, observing that most circles die and only a few survive. In Section~\ref{sec:survival-of-the-fittest}, we investigate how many circles survive and which circles survive under different circumstances. In Section~\ref{sec:circle-dynamics}, we show that the dynamics of circles can be well modeled by a linear differential equation. Related works are discussed in Section~\ref{sec:related-works}. We conclude in Section~\ref{sec:conclusions}.

\section{Problem Setup}\label{sec:problem-setup}

\begin{figure}[htb]
    \centering
    \begin{subfigure}[t]{\textwidth}
        \includegraphics[width=\linewidth]{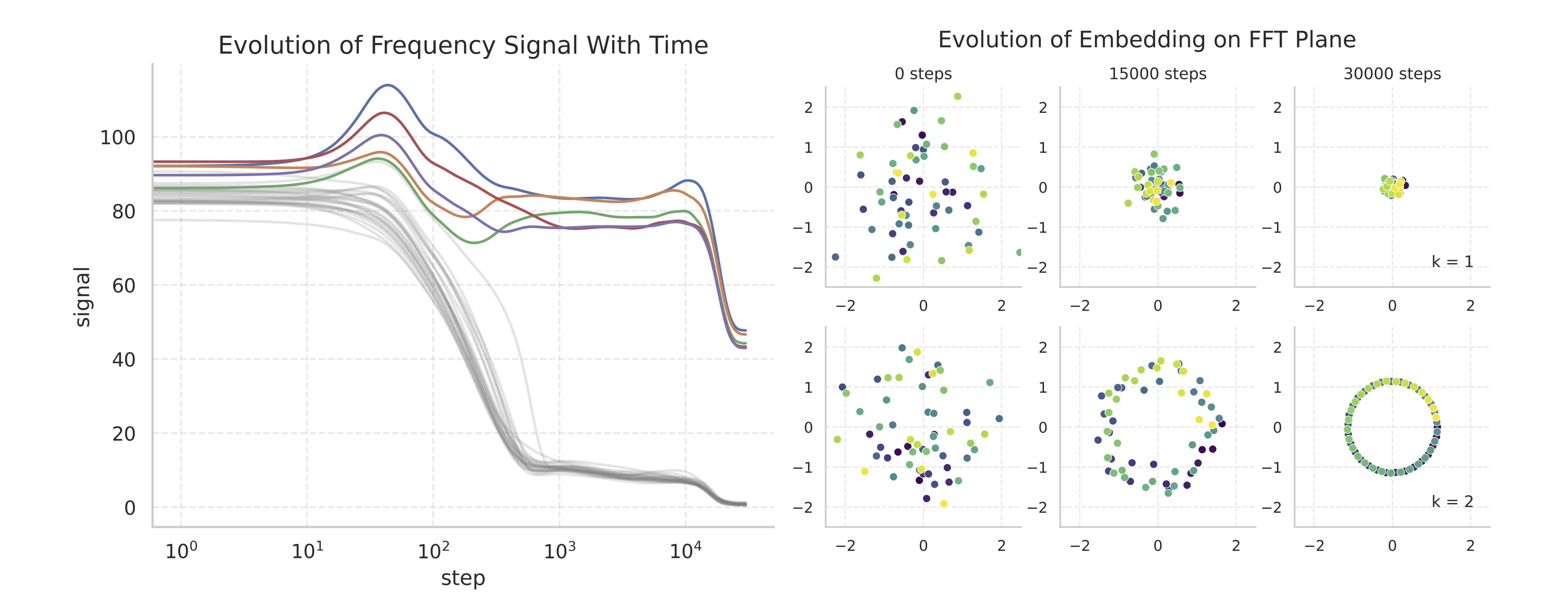}
    \end{subfigure}
    \caption{(\textit{Left}) The signals, the magnitude of the Fourier coefficients, of each embedding frequencies over time shown on a logarithmic scale. The surviving frequencies clearly separate themselves from the rest of the frequencies that quickly go to 0.  (\textit{Right}) Snapshots of the embedding projected onto a dead frequency (top) and a survived frequency (bottom) at different timesteps during training.}
    \label{fig:evolution}
\end{figure}

\label{sec:modular-addition}
\paragraph{Modular addition} We study models performing the task of modular addition in the form of $a + b = c\ ({\rm mod}\ p)$, where $a, b, c = 0, 1,\ldots, p - 1$.  Our models have an embedding matrix $\mat{W}_E$ of size $(p, d)$, where every integer $t \in \{0, 1, \cdots, p-1\}$ is treated as a token and has an associated embedding vector $\mat{E}_t \in \mathbb{R}^d$. The model tokenizes the two inputs $a$ and $b$, concatenates them, and feeds $[\mat{E}_a, \mat{E}_b] \in \mathbb{R}^{2d}$ to a two-layer MLP with two hidden layers of 100 neurons each, producing a categorical output $c$. 
We default to $d = 128$ and a large weight decay of $0.5$ to make sure the model quickly "groks" \citep{power2022grokking} to form the final representation.

\begin{figure}[htb]
    \centering
    \begin{subfigure}[t]{\textwidth}
    
    \includegraphics[width=\linewidth]{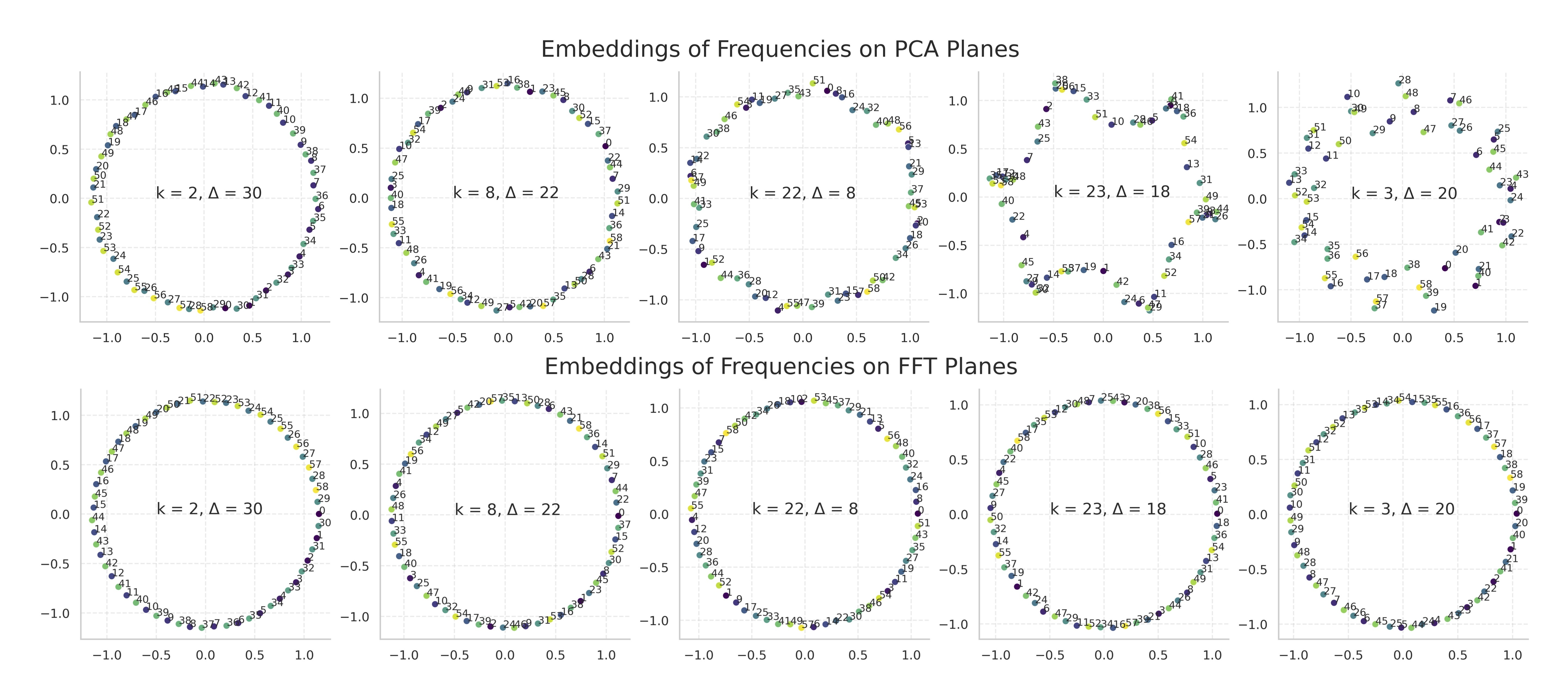}
    \end{subfigure}

    \caption{Top: The model embedding projected onto the first 10 principal components in pairs, the only components with significant singular values. Bottom: The model embedding projected onto the Fourier basis of frequencies in descending order of signal magnitude. The $\Delta$ between adjacent tokens shows a correspondence between PCA and FFT.\protect\footnotemark This indicates that PCA is a loose approximation of circles associated with Fourier frequencies.}
    \label{fig:fourier_basis}
\end{figure}

\label{sec:circles-and-signals}
\paragraph{Circles and signals} Prior work has shown that circular representations (circles) are important for neural networks to perform modular addition~\cite{liu2022towards,nanda2023progress,zhong2023the}. However, as shown in Figure \ref{fig:fourier_basis}, neighboring numbers along the circle may have increments other than 1 because there are $p$ equivalent group representations that correspond to circles with different Fourier frequencies $k=1,2,\cdots,(p-1)/2$.~\footnote{Note that frequency $k$ and $p-k$ refer to the same circle, which accounts for the factor 2.} The circle of frequency $k$ places a token $t$ at $({\rm cos}(2\pi kt/p),{\rm sin}(2\pi kt/p))$.

Similar to~\citet{nanda2023progress}, we decompose representations into a linear combination of circles of different frequencies. Denoting $\mat{E}_n\in\mathbb{R}^d$ to be the embedding vector for token $n$, we define the Fourier coefficients of frequency $k$ to be
\begin{equation}
    \mat{F}_k = \sum_{n = 0}^{p - 1} e^{-i 2 \pi \frac{k}{p} n}\mat{E}_n.
\label{eq:fourier-transform}
\end{equation}

The circle of frequency $k$ is located on the plane spanned by two vectors $\mathrm{Real}(\mat{E}_k), \mathrm{Imag}(\mat{E}_k) \in \mathbb{R}^d$. %
We define the Fourier signal of frequency $k$ as $\|\mat{F}_k\|^2 = \sum_{j = 0}^{d - 1} (\mat{F}_k^j)^2$ and 
observe how the signal of each frequency evolves over training steps, shown in Figure \ref{fig:evolution}(Left). We observe that only a few circles have significant signals (hence \textit{survived}) in the end, while the signals of all other circles decay to almost zero (hence \textit{dead}). Shown in Figure \ref{fig:evolution}(Right), projections of a surviving frequency stabilize into a clear circle (bottom), while a dead frequency collapses towards its center (top). While prior works have found circles either with Fourier transformation \cite{nanda2023progress} or with principal component analysis (PCA) \cite{liu2022towards}, we use the Fourier transformation since it can reveal circular representations better (see Figure~\ref{fig:fourier_basis}). Because each frequency corresponds to a circle, we use the phrase ``circle'', ``circular representations'' or ``frequencies'' interchangeably throughout the whole paper.

\label{sec:research-question}
\paragraph{Research question: which circles survive?} In Figure~\ref{fig:evolution}(Left), we observe that the surviving circles in the end tend to have higher initial signal. Can we use this information to predict survived circles? Unfortunately, we find that initially large frequencies do not \textit{always} lead to surviving circles, but large initial signals do lead to \textit{higher probability} of surviving. 
Therefore, we resort to statistical analysis rather than deterministic analysis by aggregating training results with a fixed embedding and random initializations of MLPs and datasets and report the mean and 95\% confidence interval. %

\section{Survival of the Fittest}\label{sec:survival-of-the-fittest}
In an ecological system with constrained resources, only the fittest will survive. We hypothesize that this theory also holds true in the case of modular addition. Motivated by the ecological analogy, we wish to analyze the competitive dynamics of how the model selects certain circles as its representation by answering 
{\bf Q1}: how many circles survive? {\bf Q2}: what are the properties of the circles that survive? 

\subsection{{\bf Q1}: How many circles survive? A Resource Perspective.}\label{sec:how-many}

\footnotetext{$\Delta$ can be calculated as the inverse of frequency $k$ modulo $p$ so that $k \cdot \Delta = 1  (\mod p)$.}

In ecological systems, with more resources available, it is intuitive to think that more species could have a chance to survive. Similarly, in neural networks, the embedding dimension can be made analogous to the total resources available, as a larger model has stronger approximation power and can lead to better performance on the desired task~\cite{sharma2020neural,michaud2024quantization,song2024resource}. We expect that higher-dimensional embeddings, even just randomly initialized, provide more ``resources'' for model training. Specifically, in the setting of modular addition, when $d \geq p$, with probability one, perfect circles can be obtained by linearly projecting $d$-dimensional random representations into suitable subspaces.~\footnote{Finding a subspace where a perfect circle of frequency $k$ lives on is equivalent to find two linear projections $\mat{p}_1$ and $\mat{p}_2$ such that $\mat{E}_a\cdot \mat{p}_1={\rm sin}(2\pi ak/p)$ and $\mat{E}_a\cdot \mat{p}_2={\rm cos}(2\pi ak/p)$ for all $a=0,1,\cdots,p-1$. Since these equations are linear and linearly independent (due to random initializations), when the number of unknown variables $2d$ is larger (smaller) than the number of equations $2p$, the system is underdetermined (overdetermined), leading to the existence (nonexistence) of solutions.} 

\begin{figure}[htb]
    \centering
    \begin{subfigure}[t]{\textwidth}
    
    \includegraphics[width=\linewidth]{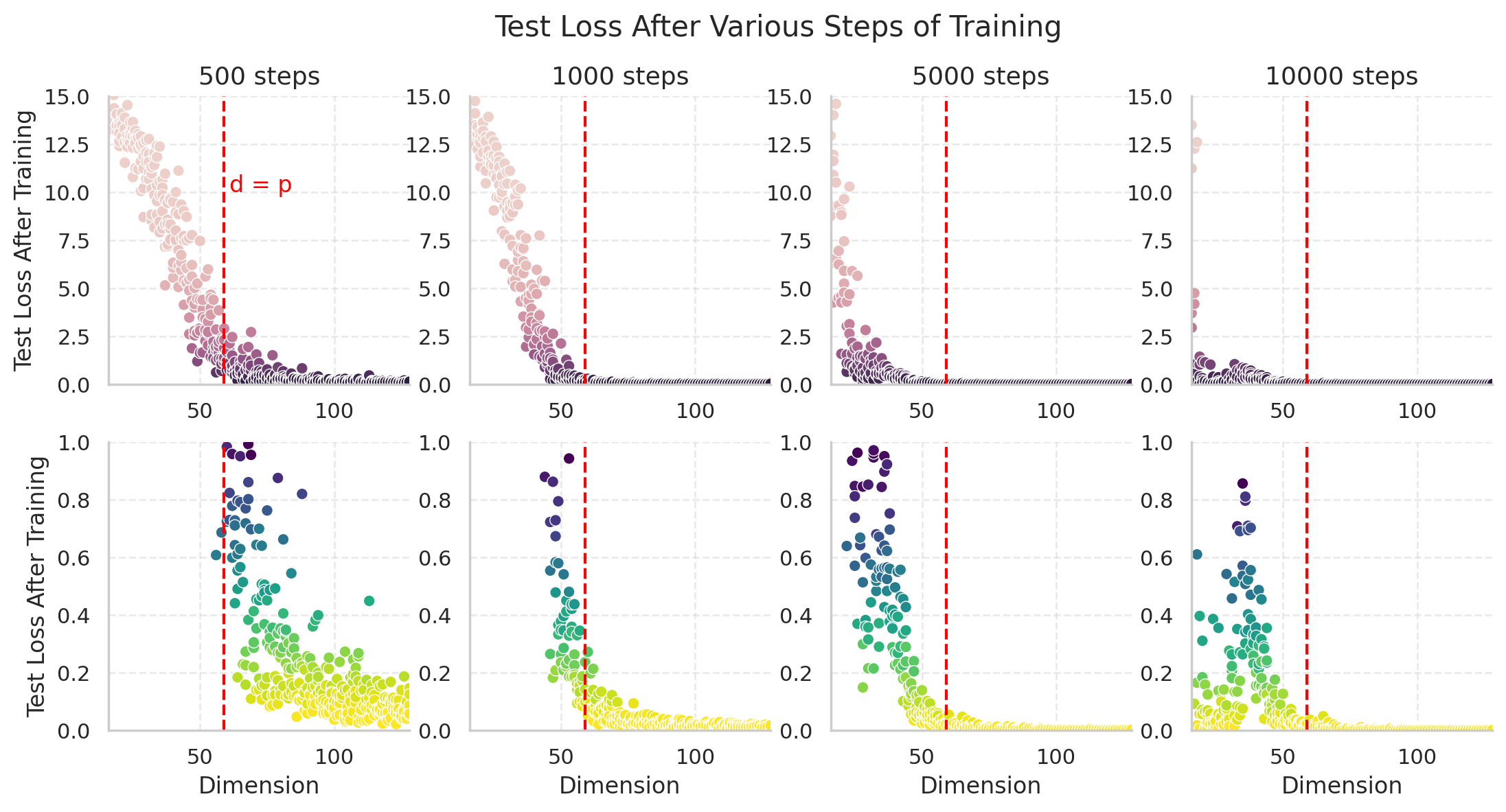}
    \end{subfigure}
    \caption{Freezing the initial embedding and training only the MLP, test loss (zoomed in on the \textit{bottom} to $< 1.0$) in relation to embedding dimension $d$. One can notice a dip in loss at $d=p$.}
    \label{fig:loss_freeze}
\end{figure}

\paragraph{Freezing embeddings} \label{sec:freeze-embed} We confirm with experiments that the initial random embeddings already encapsulate rich representations for MLPs to learn. To verify, we freeze the embeddings at their initialization and train only the MLP. We vary the dimension $d$ of the embedding and observe evolution of test loss. In Figure \ref{fig:loss_freeze}, we show snapshots of the test loss at different timesteps as a function of embedding dimension, where we indeed notice a phase transition around $d = p$---the test loss is sharply better when $d > p$ than when $d < p$.
Interestingly, we observe that at 10,000 steps, models with the small $d$s ($d \leq 20$) have lower loss than models with ``medium" $d$s ($20\leq d\leq 59$), an intriguing observation similar to the phenomenon of double descent \citep{Nakkiran2020Deep}. Our speculation is: networks with small $d$s can benefit from feature learning, while networks with large $d$s ($d \geq p=59$) have more sheer approximation resources, as we argue above, despite being in a lazy learning regime \citep{Geiger_2020}. Networks of medium $d$s fail perhaps because they are both lazy and resource constrained. A full investigation of this phenomenon is left for future work.

\begin{figure}[b]
    \centering
    \begin{subfigure}[t]{0.47\textwidth}
        \includegraphics[width=\linewidth]{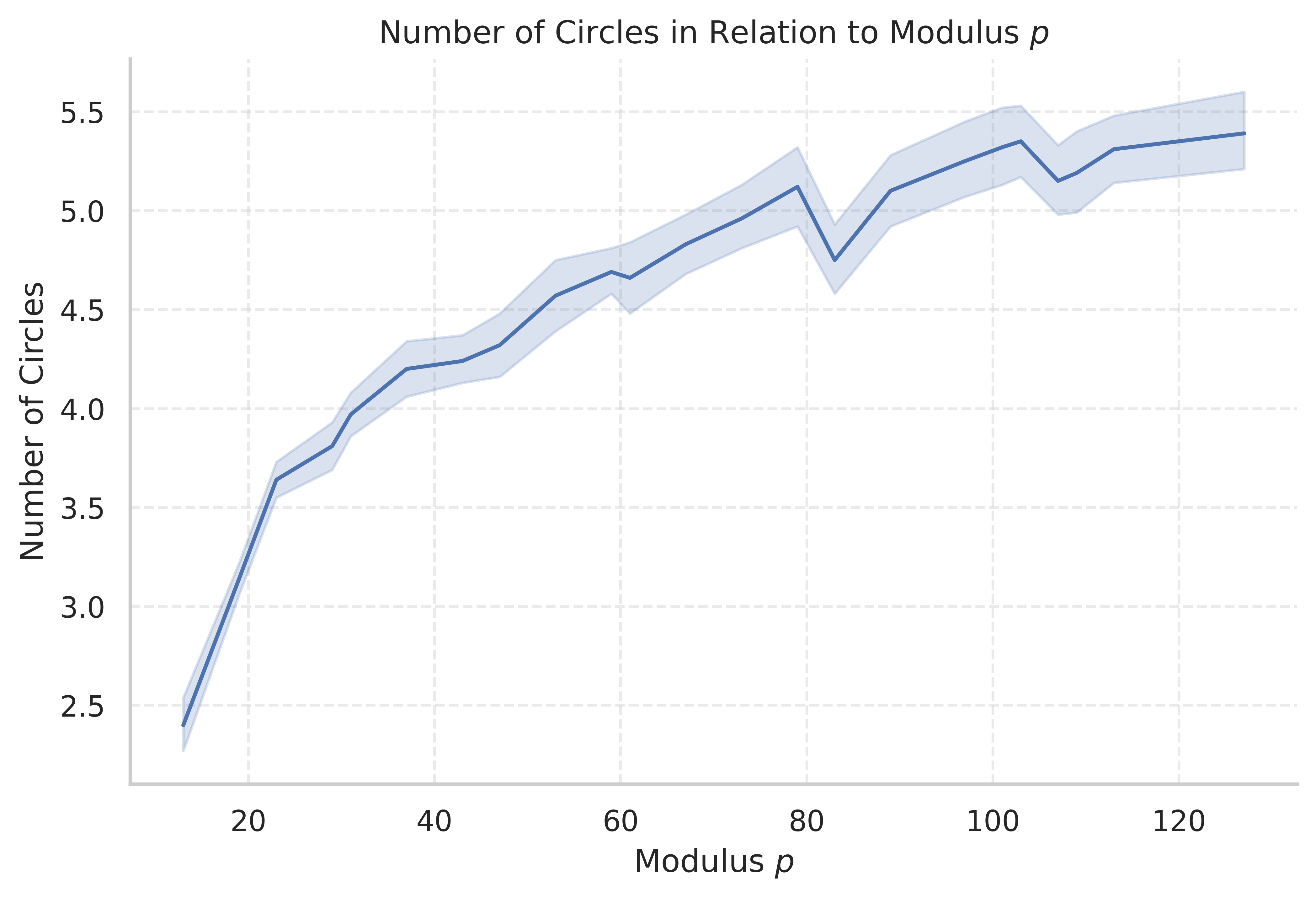}
        \label{fig:num_circles_prime_mod}
    \end{subfigure}
    \begin{subfigure}[t]{0.47\textwidth}
        \includegraphics[width=\textwidth]{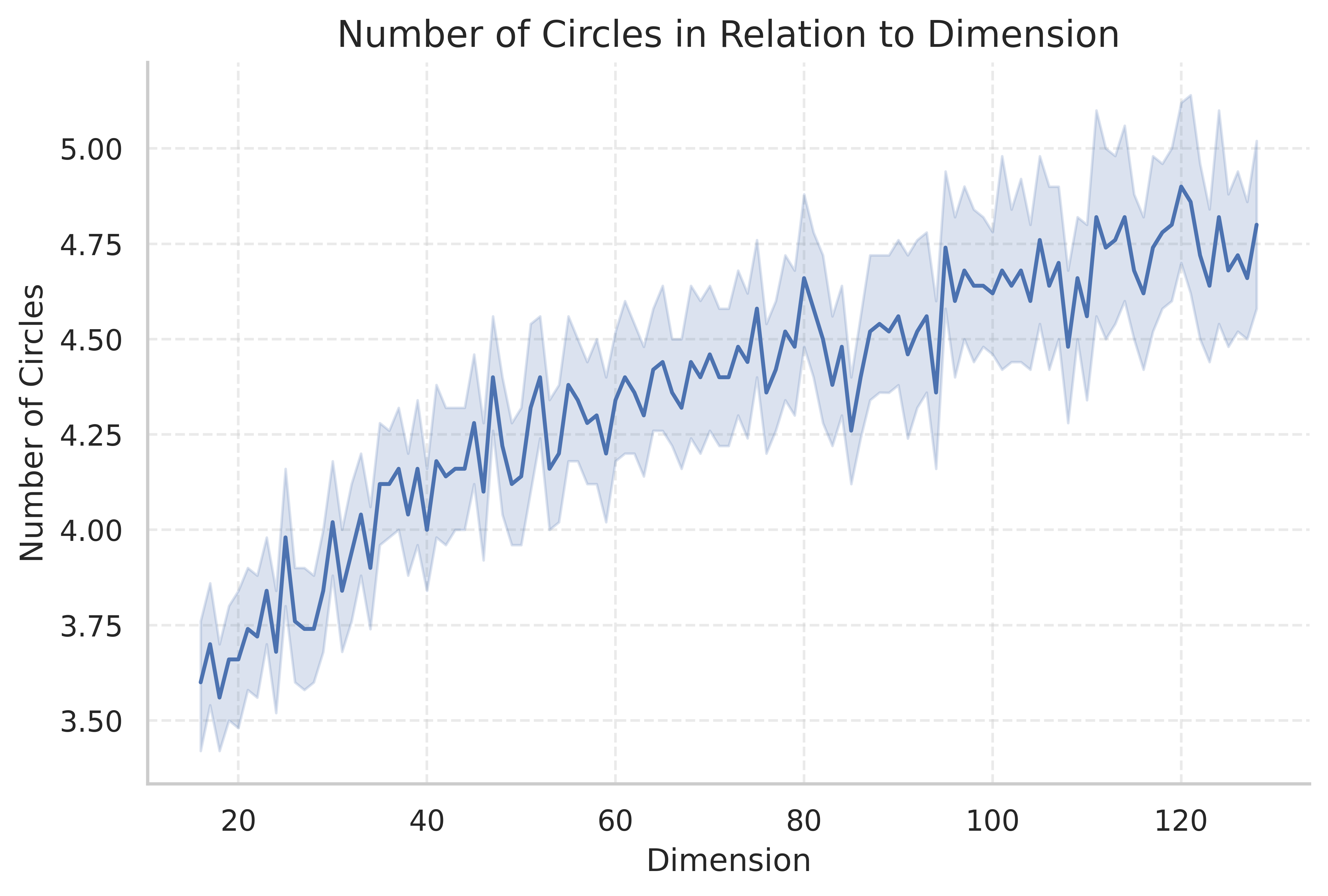}
        \label{fig:num_circles_dim}
    \end{subfigure}
    \caption{(\textit{Left}): The number of circles the model chooses for its final representation in relation to the number of tokens, $p$, over 100 random trials. (\textit{Right}): The number of circles as the embedding dimension for representing each token increases from 16 to 128. %
    }
\label{fig:num_circles}
\end{figure}

\paragraph{Trainable embeddings} \label{sec:trainable-embed} Now, we go back to the standard setup where embeddings are trainable. Since dimensionality is analogous to resources in ecological systems, we want to understand if more circles can survive in embeddings of larger dimensions. The answer is yes. 
As both $p$ and $d$ determine the embedding dimension, we conduct experiments varying one while fixing the other. In Figure \ref{fig:num_circles}(Left), we fix $d$ and study the effect of varying $p$ on the expected number of surviving frequencies after sampling over 100 trials, from which we identify clear positive correlation between $p$ and the number of circles the model learns.
Similarly, we fix $p$ and vary $d$ to see $d$'s effect on the total number of surviving frequencies in Figure \ref{fig:num_circles}(Right), which similarly shows an upward curve. As the embedding dimension increases, the number of circles (algorithm redundancy) increases. This redundant mechanism potentially makes neural networks more robust, but less parameter-efficient and interpretable. Understanding this mechanism would be an intriguing topic for future work.

\subsection{Q2: Which Circles? Some Are ``Fitter" at Initialization.}
\label{sec:which-circles}

We want to predict the final surviving circles from the initialization. Following the \emph{Survival of the Fittest hypothesis}, we wish to define a few ``fitness'' measures that make certain frequencies more likely to survive than others.
We indeed observe some properties that make some frequencies more ``fit,'' including large initial signals and large initial gradients. Intuitively, one can think of these ``fitness'' measures as being born either strong or fast in the ecosystem analogy.

\begin{figure}[htb]
    \centering
   \begin{subfigure}[t]{\textwidth}
        \includegraphics[width=\linewidth]{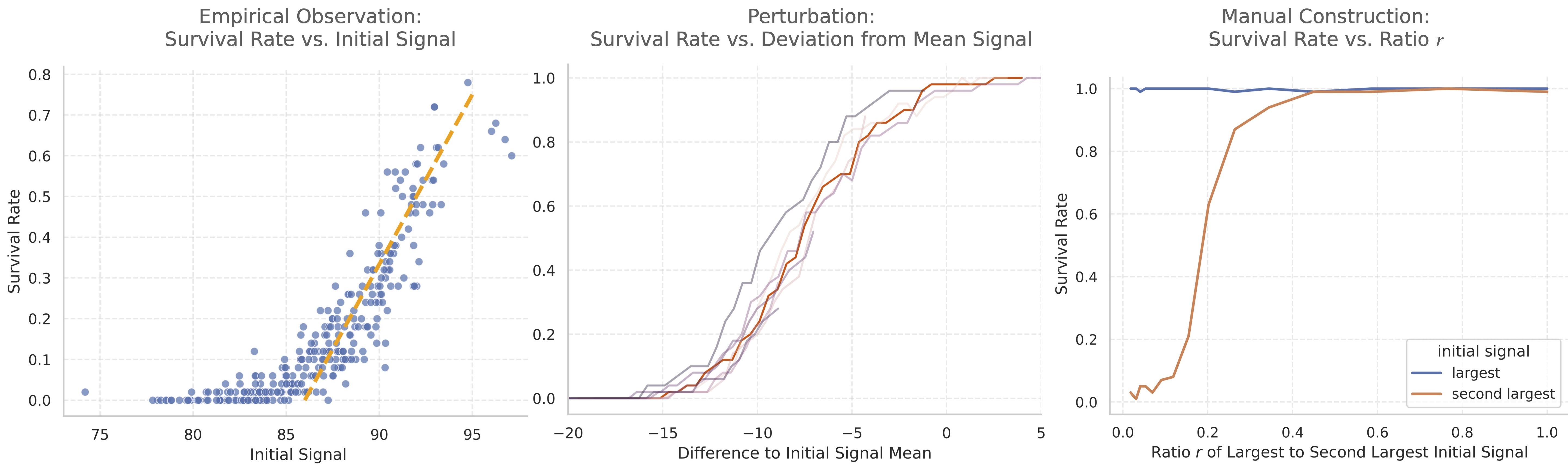}
    \end{subfigure}
    \caption{(\textit{Left}) Survival rates of frequencies given their initial Fourier signals over many randomized trials. (\textit{Middle}) Survival rate of an arbitrary perturbed frequency versus deviation from the mean initial signal, with different lines showing the same frequency in different embeddings. (\textit{Right}) Survival rates of the largest frequency and the \nth{2} largest as they differ by $r$. }
\label{fig:init_signal}
\end{figure}

\subsubsection{Initial Signal---\texorpdfstring{\nth{1}}{1st} Fitness Measure}\label{sec:init-signal}

We define survival rate as the number of times a particular frequency survives and becomes part of the final representation over the total number of the trials where we fix the embedding and randomize over different MLPs and dataset initializations. In Figure \ref{fig:init_signal}(Left), we show that over 10 different embedding initializations and 50 trials each, establishing a linear correlation: the higher the initial signal, the more likely the model is to choose that frequency as its final representation. To confirm, we find the Pearson correlation between survival rate and initial signal is $0.85$ with a p-value smaller than $10^{-3}$.  

To corroborate, we conduct a perturbation experiment on the initial embedding. Specifically, for a given embedding at initialization, we perform Fourier transform and find the initial coefficients of an arbitrary frequency. We manually enlarge or shrink its magnitude and perform an inverse FFT to restore the embedding, from which we perform model training. We demonstrate in Figure \ref{fig:init_signal}(Middle) the survival rate of the perturbed frequency in different initializations as it deviates from the mean of the rest of the signals. We observe that if the frequency is much higher than the rest, survival rate is near 100\%, while the frequency rarely survives if it is much lower than the mean. This experiment indeed suggests that as we control the environment much more closely, the initial signal of the frequency plays a unique role in determining the final representation. 

To further substantiate our findings, we manually construct an embedding to evaluate survival rate. We first randomly sample 2 out of $p$ frequencies and set their signals to be of a varying ratio $r \in [0,1]$. Concretely, the largest frequency will have signal magnitude $s$, while the second highest will have a signal $r\cdot s$. We set all other frequencies to have a signal of small $\epsilon = 10^{-6}$. In this setup, we show in Figure~\ref{fig:init_signal}(Right) that the frequency with the highest signal will always survive, while the second highest frequency increases in survival rate as its signal increases and differentiates itself further from the rest of the signals. Interestingly, despite all other frequencies having a signal near $0$ at initialization, the model sometimes chooses to revive them rather than always choosing the two clear frontrunners, a phenomenon that warrants more investigation in the future.

\subsubsection{Initial Gradient---\texorpdfstring{\nth{2}}{2nd} Fitness Measure}
\label{sec:initial-gradient}

In evolution theory, species best adapted to their environment would survive. In neural networks, we hypothesize that not only the representations with high initial signals, but also those can quickly adapts into circles are more likely to survive. This observation motivates us to examine initial expanding velocity (gradient).
We simply calculate the gradient as the difference in the signals before and after a given timestep $i$, taking into account both the embedding gradients and weight decay.

\begin{figure}[htb]
    \centering
    
    \begin{subfigure}[t]{\textwidth}
        \includegraphics[width=\linewidth]{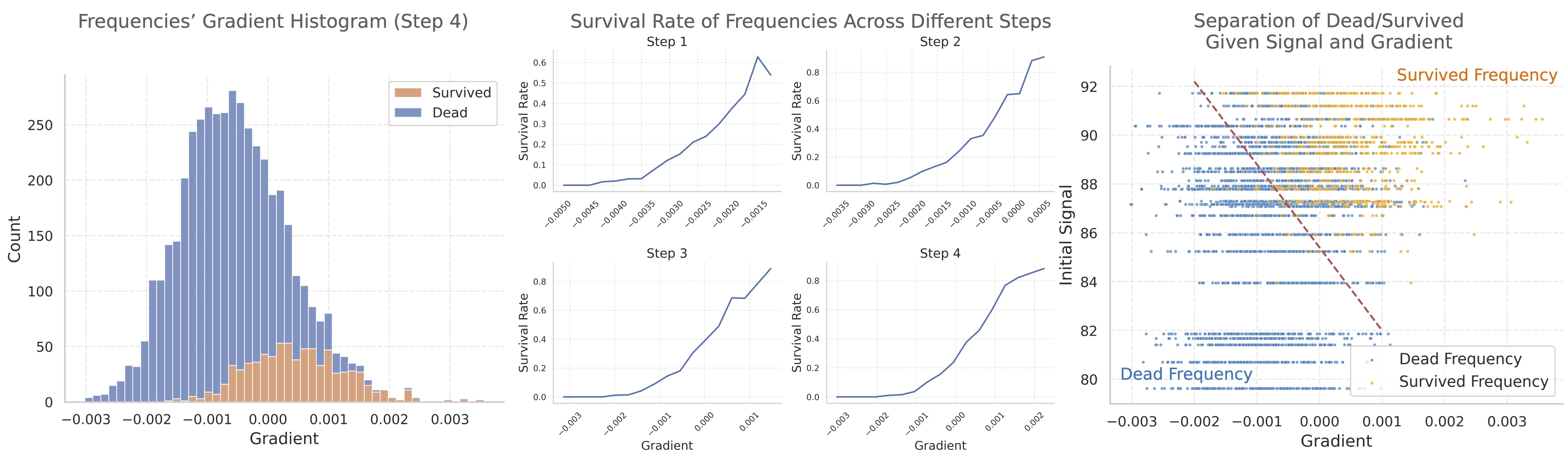}
    \end{subfigure}
    \caption{Distribution of the frequency gradient at step 4. (\textit{Middle}) Survival rate in relation to frequency gradients for step 1 to 4. (\textit{Right}) Survived (orange) and dead (blue) frequencies characterized by their initial signals and gradients at step 4. }
\label{fig:init_gradient}
\end{figure}

In Figure \ref{fig:init_gradient}(Left), we show the frequencies' initial gradient values alongside their survival status. Due to the weight decay mechanism, all gradient signals decrease over time, but those with higher initial gradients tend to shrink less and are more likely to survive. In Figure \ref{fig:init_gradient}(Middle), we show that as the initial gradient increases, the survival rate of those frequencies increases. To analyze the possibly compounding effect, we show that frequencies with both high initial signal and gradient are more likely to survive in Figure \ref{fig:init_gradient}(Right), as the top right corner is more lit with oranges, indicating more survived frequencies. 
To further verify the above observation, we train a linear support vector machine (SVM) that separates the dead and survived frequencies, achieving an $83.8\%$ accuracy.

\paragraph{Relation to Lottery Ticket Hypothesis} \label{sec:lottery-ticket-relation} Our analysis can be related to the Lottery Tickets Hypothesis (LTH)~\cite{frankle2018the}, where some subnetworks are ``winning tickets`` that achieve comparable test accuracy to the original network when trained in isolation. Our results suggest that with large embedding dimensions, good circles exist even at initialization, similar to ``winning tickets.'' Our analysis is technically different from LTH in two ways: (1) LTH requires training and pruning to identify ``winning''  tickets, while circles are mathematically defined without training (but specific to modular addition); (2) LTH only states the existence of ``winning''  tickets at initialization, while we manage to characterize the properties of ``winning'' circles, that, when trained in isolation, have comparable accuracy (see Appendix \ref{appendix:two_circles}). Our work provides representation-level insights to the studies of LTH.

\section{Circles Can Collaborate or Compete}\label{sec:circle-dynamics}

In the last section, we demonstrated how \textit{Survival of the Fittest} explains the evolution of circles. However, what is lacking from this explanation is the interaction between circles, considering ``fitness'' is defined on individual circles. This section seeks to understand circle interaction from two aspects: (1) understand how a group of $n$ circles ``collaborate'' to reduce losses on the task at hand; (2) understand the (effective) differential equations that govern the evolution of circle signals. %

\subsection{Circles Have to Collaborate to Reduce Loss}\label{sec:circle-collaborate}

\begin{figure}[b]
    \centering
    \begin{subfigure}[t]{\textwidth}
        \includegraphics[width=\linewidth]{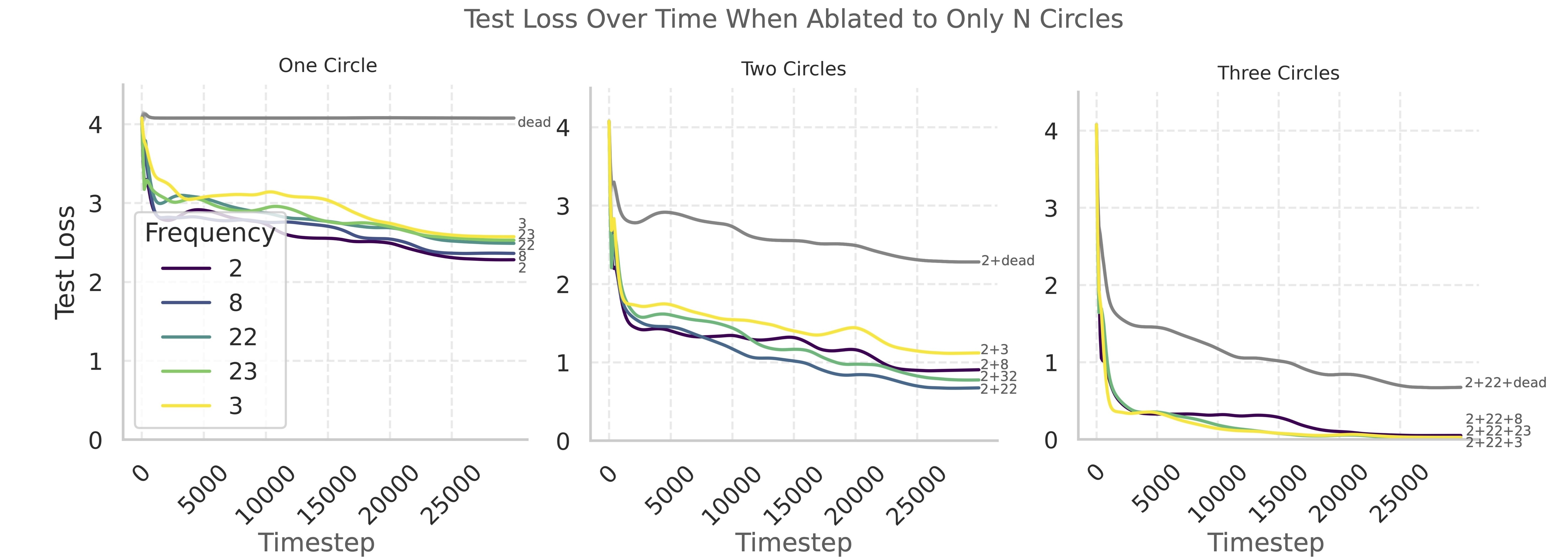}
    \end{subfigure}
    \caption{Test loss over training steps in ablation study, with only one (left), two (middle), three (right) circles remaining andother circles removed. Three circles are needed to reduce the loss to zero.}
    \label{fig:interactions}
\end{figure}

We observed that models naturally form multiple circles in training. Why is this the case? Can't a single circle perform modular addition effectively? Here, we show that a strong cooperative pattern exists between the circles of different frequencies. We investigate this relationship by conducting ablation studies, by manually isolating 1, 2, and 3 different frequencies (removing all other frequencies) and see if these isolated frequencies can solve the task of modular addition. We find that the model is not able to perform the modular addition with only one circle, still has considerably high loss using two circles, and reaches near zero loss with 3 circles. The loss achieved over time in $30,000$ training steps, with 1, 2 and 3 circles respectively, is shown in Figure \ref{fig:interactions}.

\subsection{Modeling Circle Dynamics with Differential Equations}\label{sec:circle-ode}

In ecology, the Lotka-Volterra equations are famous for using first-order nonlinear differential equations to model the relationship between prey $x$ and predators $y$ \citep{alon2019introduction}. They have the following form $\mathrm{d}x/\mathrm{d}t = \alpha x - \beta xy, \mathrm{d}y/\mathrm{d}t = \delta xy  - \gamma y$, where $x$ and $y$ represent the population density of prey and predators, respectively, and $\mathrm{d}x/\mathrm{d}t$ and $\mathrm{d}y/\mathrm{d}t$ are the instantaneous growth rates of the two populations. This can be easily generalized to more than two species by involving linear single-body terms and quadratic two-body terms. Interpreting species population as frequency signals, we have:

\begin{equation}
\dfrac{\mathrm{d}x_i}{\mathrm{d}t} = \sum_{i=1}^{N_c} \alpha_i x_i + \sum_{i=1}^{N_c}\sum_{j=i}^{N_c} \beta_{i,j} x_i x_j, \quad i = 1, 2, \cdots N_c, \label{eq:second_order}
\end{equation}
where $x_i$ represents signals of each frequency.

Although the fit yields an $R^2$ value close to 1, the model suffers from overfitting due to too many free parameters of $x_i \cdot x_j$ terms. When we attempt to compute a trajectory using our estimates, errors accumulate, leading to a rapid loss of numerical stability. This prompts a natural question: are the quadratic nonlinear terms really necessary? 

\begin{figure}
    \centering
    \begin{subfigure}[t]{\textwidth}
        \includegraphics[width=\linewidth]{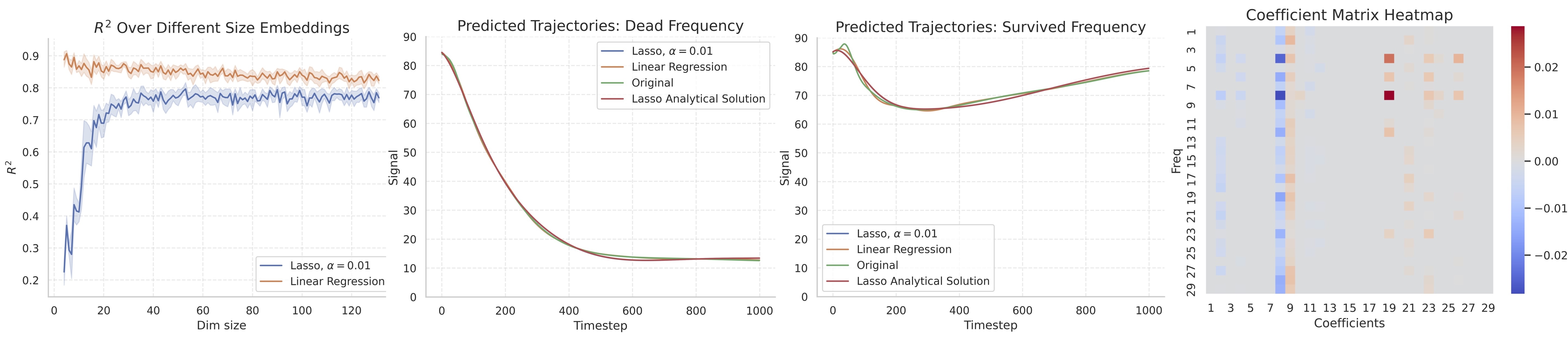}
    \end{subfigure}
    \begin{subfigure}[t]{0.22\textwidth}
        \captionsetup{skip=0pt}
        \caption{$R^2$ for fit on embeddings of varying $d$}
        \label{fig:ode_R^2}
    \end{subfigure}
    \hspace{0.1cm}
    \begin{subfigure}[t]{0.22\textwidth}
        \captionsetup{skip=0pt}
        \caption{Estimated Trajectory for Dead Frequency}
        \label{fig:ode_dead_traj}
    \end{subfigure}
    \hspace{0.1cm}
    \begin{subfigure}[t]{0.22\textwidth}
        \captionsetup{skip=0pt}
        \caption{Estimated Trajectory for Surviving Frequency}
        \label{fig:ode_survived_traj}
    \end{subfigure}
    \hspace{0.1cm}
    \begin{subfigure}[t]{0.22\textwidth}
        \captionsetup{skip=0pt}
        \caption{Coefficient Matrix for Lasso regression}
        \label{fig:ode_coef_matrix}
    \end{subfigure}
    \caption{Using linear ODE to model circle interactions during training.}
\label{fig:ode}
\end{figure}

To our satisfaction, the answer is no. Deviating from the Lokta-Volterra equations, we found an even simpler, linear differential equation, that can estimate the training trajectory well. Removing the second-order terms from Equation \ref{eq:second_order}, we get 
\begin{equation}\label{eq:linear}
\dfrac{\mathrm{d}x_i}{\mathrm{d}t} = \sum_{j=1}^{N_c} \alpha_{i,j} x_j+b_i,\quad i = 1, 2, \cdots, N_c,
\end{equation}
or in matrix form $\frac{\mathrm{d}\mat{x}}{dt} = \mat{A}\mat{x} + \mat{b}$. With this new set of equations, we can model the trajectory well. We report the $R^2$ of our fit for both linear regression and Lasso in Figure \ref{fig:ode} over many different trials with embeddings of varying size. For two frequencies in a given trained model, one survived and one dead, we compare the original trajectory, the estimated trajectory from Linear Regression and from Lasso in Figure \ref{fig:ode}. While linear regression gives us an almost perfect fit at the risk of overfitting, Lasso provides comparable estimations with reasonable errors bounds and extremely sparse coefficients.

As the differential equations are linear, we can find an analytical solution to the ODE system. Assume $\mat{A}$ as the coefficient matrix of the regression model and $\mat{b}$ as the intercept, for a given $\mat{x}_0$, we have the following solution
\begin{equation}
    \mat{x}(t) = e^{\mat{A}t}\mat{x}_0 + (e^{\mat{A}t}- \mathit{I})\mat{A}^{-1}\mat{b}.
\end{equation}
Note that for the coefficient matrix of the Lasso fit, the matrix is extremely sparse and is not naturally invertable, so we have added a small $\epsilon \mathit{I} (\epsilon=10^{-8})$ to $\mat{A}$.
The analytical solution similarly provides us with a comparable estimation of the trajectory in Figure \ref{fig:ode}.

\paragraph{Relation to Neural Tangent Kernel}\label{sec:neural-tangent-kernel}
Despite the fact that the Lotka-Volterra equations work well to model ecological dynamics, it is not too surprising that a linear ODE is sufficient to model the training dynamics in neural networks, when the neural networks are wide enough to be  characterized by the neural tangent kernel~\cite{jacot2018neural}. However, our analysis is still novel in the sense that we can disentangle the entire embedding space into individual circles and study their interactions linearly. The idea of decomposition allows the analysis of a complicated system to be broken down to analysis of many simple subsystems and their interactions.

\section{Related Work}\label{sec:related-works}

\paragraph{Mechanistic Interpretability on Algorithmic Tasks}
A lot of work has been done to reverse engineer how neural networks implement algorithmic tasks \citep{nanda2023progress,zhong2023the,liao2023generating,chughtai2023,stander2023grokking,quirke2024understanding,quirke2024increasing} because they are mathematically well-defined and simple. However, even on these toy tasks, neural networks already display some intriguing  phenomena, including phase changes during training~\cite{nanda2023progress}, different algorithms~\cite{zhong2023the,liao2023generating}, or show the existence of multiple copies of algorithms~\cite{nanda2023progress,zhong2023the}. 

\paragraph{Training Dynamics} Training dynamics strives to understand what happens internally within a model during training. Two most studied phenomena in this area are ``grokking'' \citep{power2022grokking,liu2022towards,liu2023omnigrok,barak2023hidden,thilak2022slingshot} and ``double descent'' \citep{Nakkiran2020Deep,yilmaz,schaeffer2024double,davies2023unifying}.Other works have studied training dynamics at various abstraction levels, such as on emerging capabilities level \citep{hoogland2024developmental,mccoy2019berts}, on the circuit level \citep{olsson2022context,singh2024needs,chen2023sudden}, and on the neuron level \citep{quirke2023training}.
Similar to our analysis of circle interactions using ODEs, previous work attempted to model representation dynamics during training using simple effective dynamics~\cite{liu2022towards,baek2024geneft,van2024representations,hu2023latent}. 

\paragraph{Representation Learning} Representation learning is key for networks to generalize~\cite{huh2024platonic,bengio2013representation,le2020contrastive,zou2023representation}. Many learning paradigms aim to encourage better representations, including weak supervised learning~\cite{zhou2018brief}, contrastive learning~\cite{jaiswal2020survey,le2020contrastive}, and Siamese learning~\cite{grill2020bootstrap,chen2021exploring}. 
Similar to our paper revealing circles on different frequencies, prior works have shown redundant representations and algorithms in models~\cite{doimo2023redundant,song2024resource} and similarly circular representations in general language models~\citep{engels2024not}. 

\paragraph{Lottery Ticket Hypothesis}
The Lottery Ticket Hypothesis \citep{frankle2018the} posits that some subnetworks--``winning tickets''---identified at initialization and trained in isolation can match the test accuracy of the original, dense network. It has inspired extensions, such as a stronger conjecture on finding such subnetworks without training \citep{zhou_2019_dlt,Ramanujan,malach20a,cunha2022proving,Orseau,Pensia,diffenderfer2021multiprize}, transferring winning tickets across setups \citep{morcos,chen2021elastic}, and improving methods of pruning to find the subnetworks \citep{frankle2020stabilizing,lee2018snip,frankle2021pruning,Wang2020Picking,Tanaka}.
Our work relates to LTH that circles can be treated as subnetworks with distinct signals at initialization. 
In our setting, ``winning tickets'' exist at initialization and have nice properties like high initial signals, which allow them to eventually become the learned representations. 

\section{Conclusion}\label{sec:conclusions}
In this paper, we show that the \emph{Survival of the Fittest} theory can explain the training dynamics of the toy modular addition task. Qualitatively, embeddings can be decomposed into circles of different frequencies, deemed as species interacting with one another. Under the resource constraint of model sizes, circles with large signals and gradients are more likely to survive. Quantitatively, the dynamics of circle interaction can be described by a simple linear differential equation. Our results highlight simple laws underlying seemingly complicated representation dynamics and open the door for more fine-grained analysis of representation dynamics for mechanistic interpretability. 

\paragraph{Limitations} \label{sec:limitations} We have focused on a single learning problem: modular addition. 
Our work studying dynamics between different representations is made possible because representations in modular addition models are well-defined and well-understood.
Significant additional work is needed to scale these analyses to even more complex, general models, which remain a challenge.

\paragraph{Broader Impact} \label{sec:broader-impact} 
Understanding training dynamics allows us to better understand and control neural networks in ways we desire, such as making them more accurate and safe. Any dual-use technologies have accompanying risks, so one should exercise caution when deploying these techniques.

\section*{Acknowledgements}
We thank Wes Gurnee, Julian Yocum, and Ziqian Zhong for helpful conversations and suggestions. This work is supported by the Rothberg Family Fund for Cognitive Science, the NSF Graduate Research Fellowship (Grant No.
2141064), and IAIFI through NSF grant PHY-2019786.

\bibliography{refs}
\appendix

\include{appendix}

\end{document}

%% file: appendix.tex
\section{Weight Decay}
\label{appendix:weight_decay}

In the main paper, we have estimated how dimensionality is analogous to total resources in an ecosystem, while weight decay represents the resource constraint in the environment that decay at each time step.

\begin{figure}[htb]
    \centering
    \includegraphics[width=\linewidth]{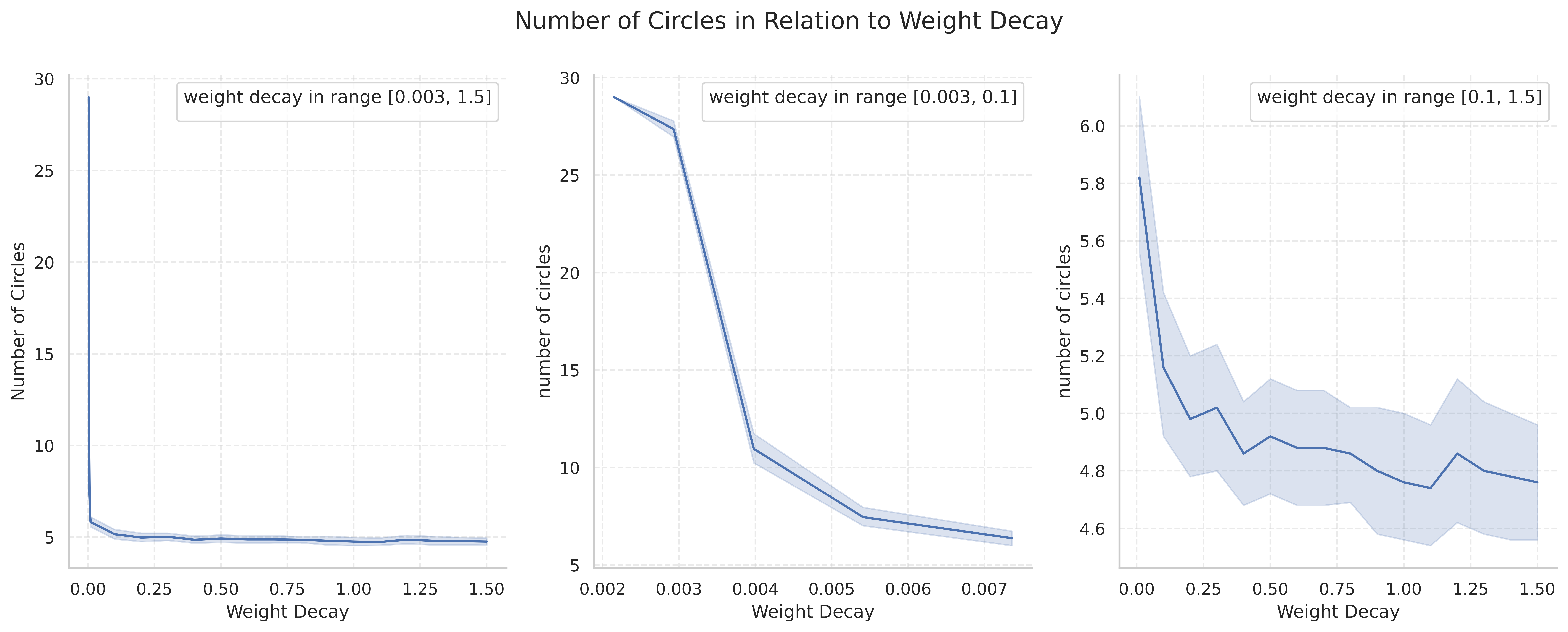}
    \caption{Number of circles survived as a function of weight decay. The \emph{left} panel displays the complete range of weight decays tested in our experiments; the \emph{middle} focuses on smaller weight decays, while the \emph{right} illustrates the transition in the number of surviving circles at larger weight decays.}
    \label{fig:weight_decay}
\end{figure}

Considering the extreme case where weight decay is zero: there is no resource limit and it is not surprising that all the frequencies survive while the neural network fails to generalize, as it has trouble "grokking" \citep{liu2022towards,liu2023omnigrok}. As weight decay slightly increases, the different frequencies pose a competitive dynamic against each other. The number of final circular representations gradually drops, as illustrated in Figure \ref{fig:weight_decay}.

\section{Circularity}
\label{appendix: circularity}

In addition to the signal magnitude metric used throughout the paper, we compute another metric, circularity, to analyze the initial embedding. We modify the metric introduced in \citet{zhong2023the} by calculating through each frequency in the Fourier Basis instead of through the principal components. As established in the paper, if $\mat{X}_0, ..., \mat{X}_{p - 1} \in \mathbb{R}^d$ are embeddings projected onto the Fourier basis, the circularity for a specific frequency $f$ is defined as

\begin{equation}
    c_k = \frac{2}{p \sum_{j = 0}^{p - 1} \mat{X}_{k, j}^2} \Bigg|\sum_{j = 0}^{p - 1} \mat{X}_{k, j} e^{2 \pi \mathbf{i} \cdot jk/p} \Bigg|^2 \label{eq:circularity}
\end{equation}

Using this metric, we conduct two experiments: one to measure the circularity of the embedding as its dimensionality varies and another to see if initial circularity plays a role in informing eventual representations. 

\begin{figure}[htb]
    \centering
     \begin{subfigure}[t]{\textwidth}
        \includegraphics[width=\linewidth]{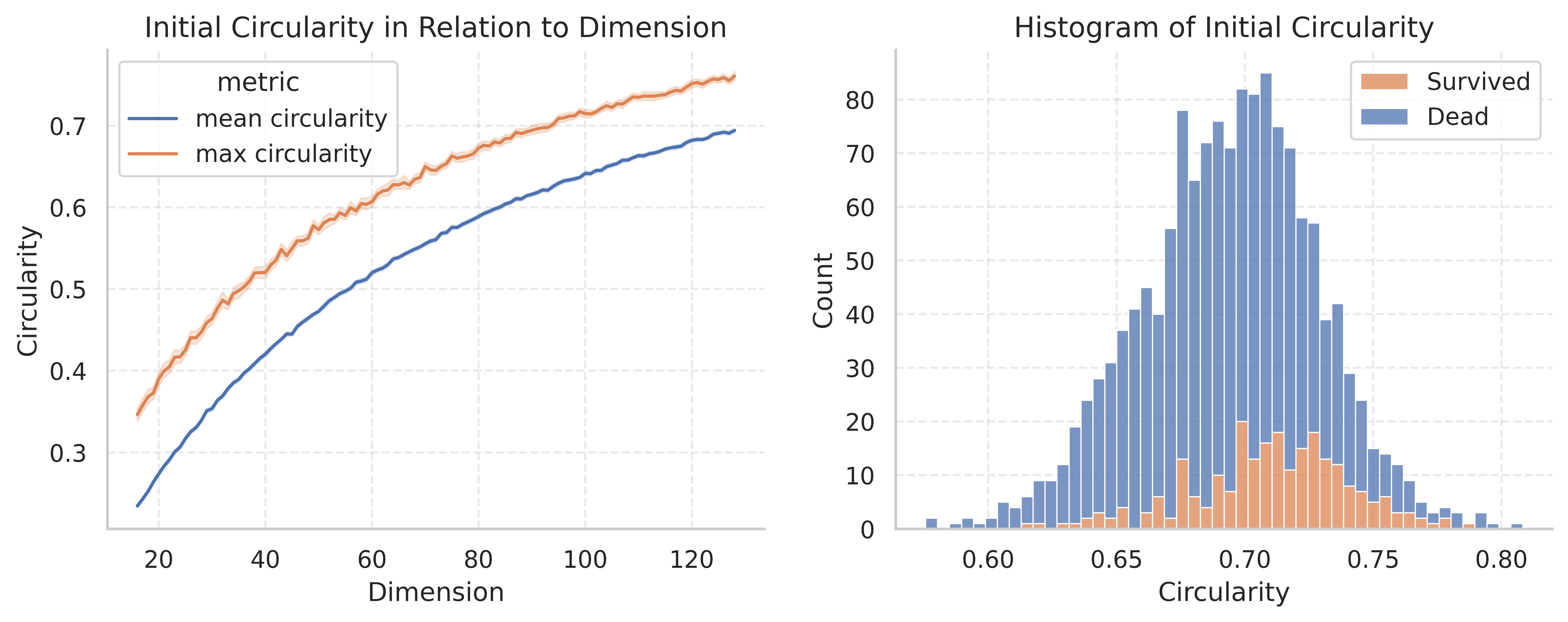}
    \end{subfigure}
    \caption{(\textit{Left}) Circularity of projections of the embedding onto different Fourier frequencies at initialization as dimension $d$ varies, calculated using Equation \ref{eq:circularity}. (\textit{Right}) Histogram of different frequencies by their gradients, along with their survival status.}
    \label{fig:circularity_dim}
\end{figure}

To investigate the impact $d$ exerts on the system, we
randomly sample 50 initial embeddings of different dimensions and compute the mean and maximum circularity among all frequencies, as shown in Figure \ref{fig:circularity_dim}(Left). Indeed, circularity increases as dimensionality increases, confirming our observation that higher-dimensional embedding encodes more complex information at initialization.

We also aim to verify the hypothesis that if the embedding is initialized closer to a circle on a given frequency, that frequency is more likely to survive.  
However, the evidence shown in Figure \ref{fig:circularity_dim}(Right) is inconclusive as to whether larger initial circularity implies a better chance of survival. 

\section{Forcing Model to Learn \texorpdfstring{$< 3$}{< 3} Circles}
In our training with an embedding of reasonably large size, such as $(p,d) =(59, 128)$, the model almost always chooses three or more circles as its final representations. As illustrated in Figure \ref{fig:interactions}, these circles cooperate to solve the modular addition task instead of acting on their own. However, one would notice that if the model were using the Clock algorithm \citep{nanda2023progress}, a single circle would be sufficient to solve the task perfectly. The observation naturally evokes the following questions: why does the model choose to learn multiple circles, and can we 'force' it to learn fewer than three circles under some contrived conditions? 

To restrict the representations the model can learn, we manually ablate the initial embedding so that only one or two frequencies have non-zero signals on the Fourier basis.

Specifically, for a given embedding, we first train without any ablation, from which we identify the original circles the model chooses to learn. We then use $k_1$ to denote the frequency with largest signal after training and $k_2$ to denote the second largest frequency. 

Using this information, we conduct four ablation experiments:
\begin{enumerate}[A)]
    \item At initialization, project the embedding onto the Fourier basis, set all frequencies except for $k_1$ to $0$, and use inverse FFT to reconstruct the embedding. Train the model using this initial embedding. 
    \item At initialization, use the same ablation procedure as above, but suppress all other frequencies except for $k_1$ and $k_2$ to $0$. 
    \item At initialization, randomly select a frequency $k_{r,1}$ and suppress all other frequencies except for $k_{r,1}$. 
    \item At initialization, randomly select two frequencies $k_{r,1}$ and $k_{r,2}$ and suppress all other frequencies except for those two.
\end{enumerate}

\begin{figure}[htb]
    \centering
     \begin{subfigure}[t]{\textwidth}
        \includegraphics[width=\linewidth]{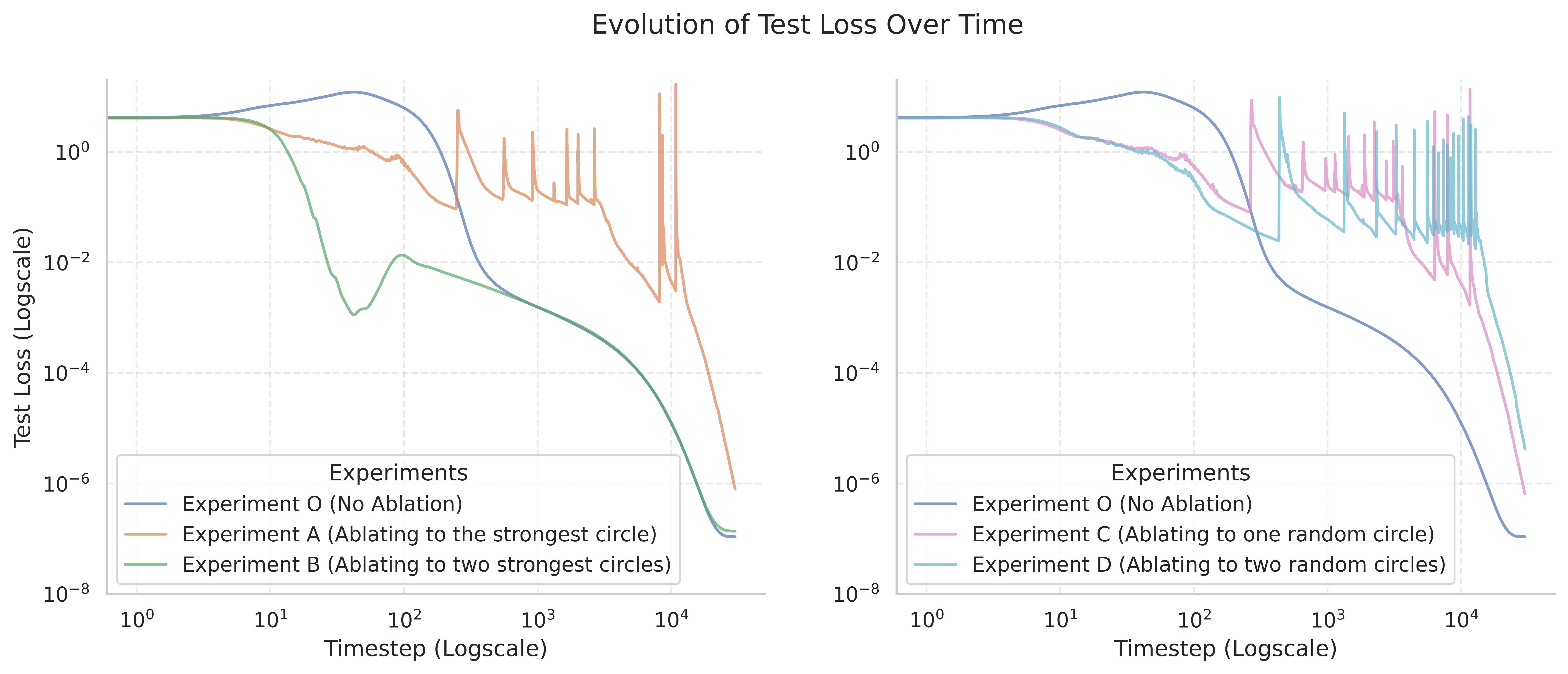}
    \end{subfigure}
    \caption{(\textit{Left}) Test loss curve over training timesteps of the ablation experiment when keeping all frequencies (blue), the largest frequency (orange), and the largest two frequencies (green). (\textit{Right}) Test loss curve of the ablation experiment when keeping all frequencies (blue), one random frequency (pink), and two random frequencies (cyan).}
    \label{fig:test_loss_evolution}
\end{figure}

In Experiment A, although the embedding is initialized with only one frequency with significant signal, the model revives some frequencies with an originally $0$ signal to form circular structures with strong signals, and eventually ends up with four learned circles. Figure \ref{fig:test_loss_evolution}(Left) shows the test loss for Experiment A in \textit{orange}. We can infer from the loss curve that despite trying to learn with only one circle, the model struggles to achieve lower loss with this simple representation and has to make its representation more complex over time, leading to the periodic spikes in test loss. In none of our experiments were we able to construct a model that naturally forms a single-circle representation.

However, in Experiment B, the model achieves good performance using only the two circles initialized with non-zero signals. Two circles seem sufficient for the model to achieve a loss as low as $1e^{-7}$, if the two circles are initalized well and we force the model to only use two. Therefore, we suspect that three circles are not necessary but rather a model choice to prefer redundant representations \citep{dalvi-etal-2020-analyzing}. Interestingly, Experiment B reaches lower test loss more quickly than the original, mainly because ablating to have only the strongest two frequencies already serves as a type of training and makes the model training process easier, similar to the findings of \citet{zhou_2019_dlt}.

In comparison, Figure \ref{fig:test_loss_evolution}(Right) shows that the model struggles to learn with only 1 or 2 random frequencies; after training, at least three circles survive. Summarizing these results, we conclude that the choice of the circles are not arbitrary: the model can learn the task well with only the two most fitted frequencies but not two random frequencies. Here, we further corroborate that the embedding initialization plays a significant role in the model's preference for its circle representations.

Note that in experiments A, C, and D, the test loss curve exhibits several cycles of spiking and subsequent decay. This slingshot phenomenon is associated with the use of the Adam optimizer and often co-occurs with grokking \citep{thilak2022slingshot}. We do not discuss this further, as it falls outside the scope of our research.

\label{appendix:two_circles}

\section{Embedding Gradients}
In Section \ref{sec:initial-gradient}, we approximate the gradient as the difference of signals before and after a given timestep. We provide further justification here.

\begin{figure}[htb]
    \centering
    \begin{subfigure}[t]{\textwidth}
        \includegraphics[width=\linewidth]{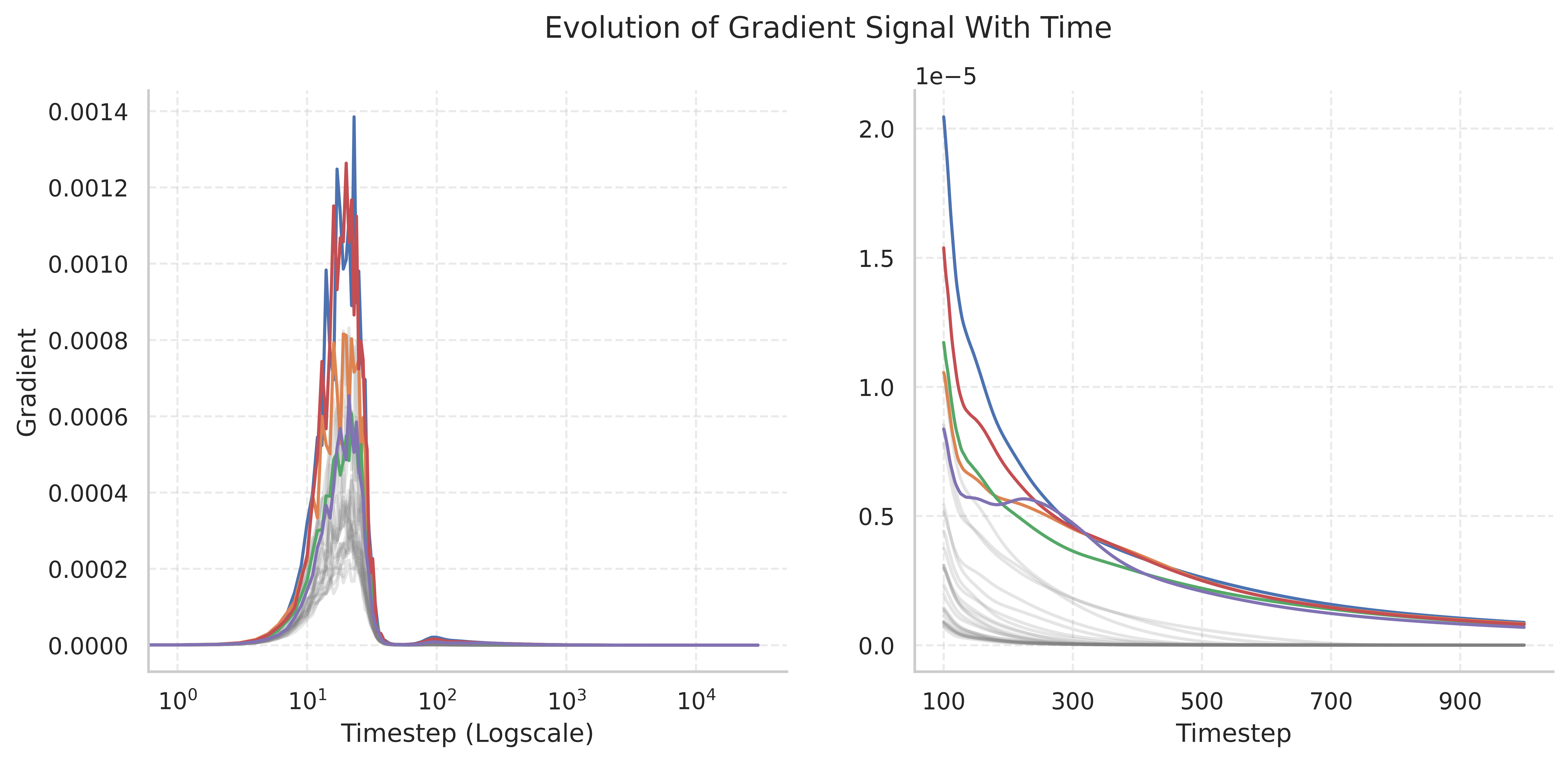}
    \end{subfigure}
    
    \caption{The embedding gradients through backpropagation projected onto the Fourier basis over time. The \textit{right} zooms into timesteps 100 to 1000. The colored curves denote frequencies that eventually survived, while the grey ones represent gradients of dead frequencies.}
    \label{fig:gradient_evolution}
\end{figure}

The actual gradient on the embeddings consists of two parts: the gradient $\frac{\partial{\mathcal{L}}}{\partial{E_k}}$ produced by MLP on the embedding through backpropagation and weight decay of the initial data.
To understand the effect of the former on frequency signals, we use the same Fourier transform procedure to transform the gradients to the Fourier basis and compute their norm, as the Fourier transform is a linear transformation.
In Figure \ref{fig:gradient_evolution}, we visualize the norm of the gradient in Fourier basis over time. The Fourier gradient spikes around 100 steps and quickly diminishes to near zero after 1000 steps. After this point, the gradient becomes negligibly small, and weight decay becomes the dominant factor affecting the evolution of signals.

These observations motivate two experimental decisions in our paper.
\begin{enumerate}
    \item Because it is difficult to reconstruct the effect of both backpropagation and weight decay compounded on top of each other on the Fourier basis, especially when weight decay dominates the gradient, we think the difference in signal is a simple and sufficient proxy to conduct experiments with in Section \ref{sec:initial-gradient}. 
    \item Since the embedding gradient becomes negligibly small after 1000 steps, we only use data from the first 1000 steps to fit the linear ODE system in Section \ref{sec:circle-dynamics}. More datapoints after the first 1000 steps will only allow the regression model to capture the dynamics of weight decay, which distracts from our study of the dynamics between representations.   
\end{enumerate}

\begin{figure}[htb]
    \centering
    \includegraphics[scale=0.55]{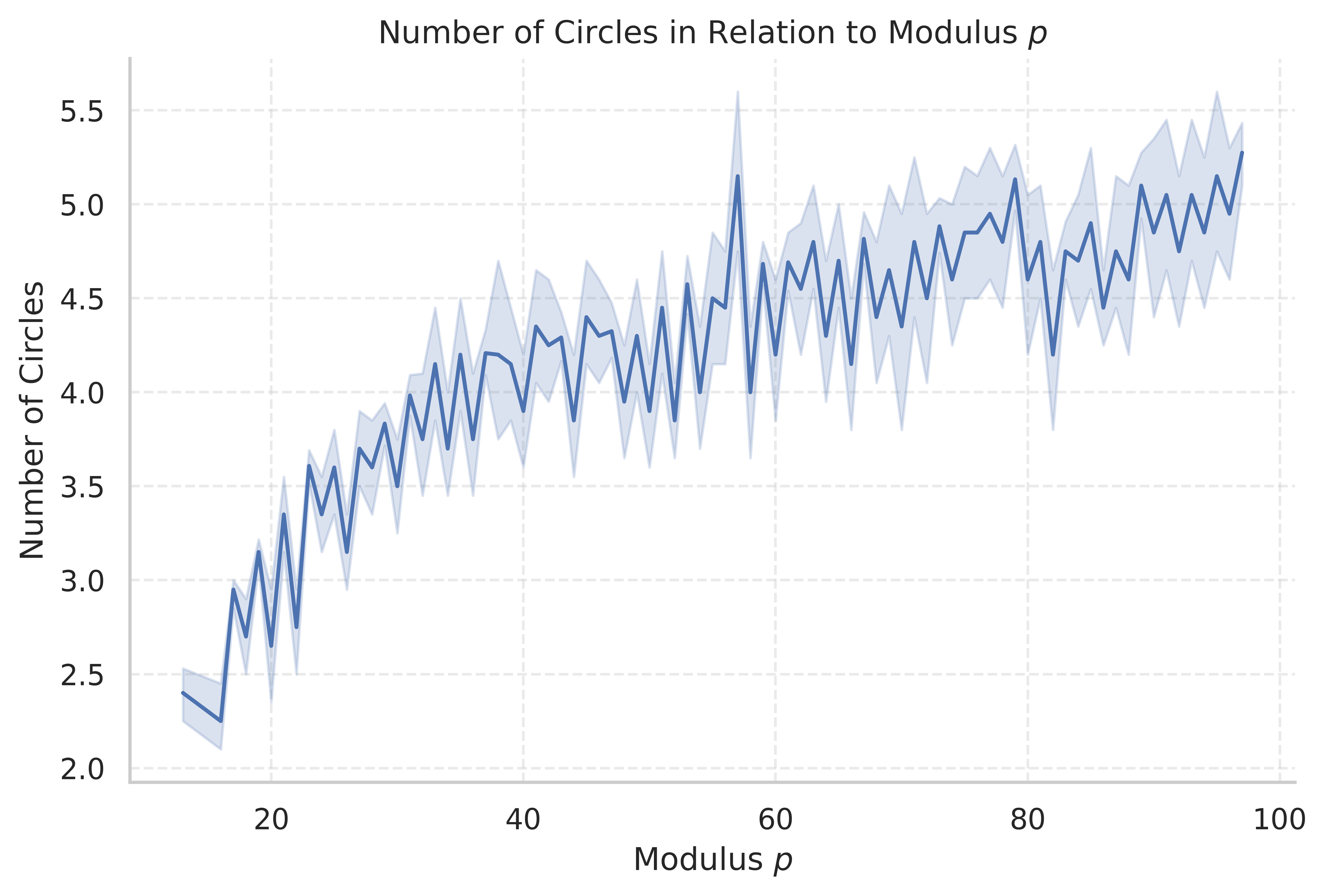}
    \caption{Number of survived circles over random trials as a function of $p$, where $p$ ranges from 17 to 97.}
    \label{fig:num_circles_mod}
\end{figure}

\label{appendix:real_gradient}

\section{Training Details}\label{appendix:training-details}

As discussed in Section \ref{sec:modular-addition}, the model we trained for the modular addition task features a simple embedding-MLP architecture. The embedding has a size of $(p, d)$, where $p$ represents the modulus and $d$ stands for the dimension. By default, we set $p = 59$ and $d = 128$. In our experiments, we vary $p$ and $d$ to study their effects on model's learned embeddings.

The MLP in our model consists of two layers: an input layer with dimension $2d$, two hidden layers, each with a width of 100, and an output layer with dimension $p$, which represents the logits for each token from $0$ to $p - 1$.

Both the embedding and the MLP are initialized from a Gaussian distribution with $\mu = 0$ and $\sigma = 1$. We train the model using the AdamW optimizer \citep{loshchilov2018decoupled}, with a learning rate set to $0.01$. The training loss is defined as the cross-entropy loss between the logits computed by the model and the ground truth. To encourage model generalization, we use a train-test split of 80-20 and apply a default weight decay of 0.5. Additionally, we experiment with other weight decay values to study their impact.

In each experiment, the model is trained for $3 \times 10^4$ steps.

\section{Non-prime Modulus \texorpdfstring{$p$}{p}}

In Section \ref{sec:freeze-embed}, we only provide results for prime modulus $p$ since non-prime $p$ behaves differently in the modular addition task due to their non-trivial factors. For example, when $p = 12$, a circle with delta $\Delta = 2$ does not cover all the numbers in $[0, p - 1]$, and our previous analysis in Fourier basis no longer holds true. However, we present supplementary results on the effect of the number of surviving circles for all possible moduli in Figure \ref{fig:num_circles_mod}.Compared with Figure \ref{fig:num_circles}(Left), one can observe more variation in Figure \ref{fig:num_circles_mod}, but an upward trend can still be observed.

\section{More ODE Approximated Trajectories}

\begin{figure}[t]
    \centering
    \includegraphics[width=\linewidth]{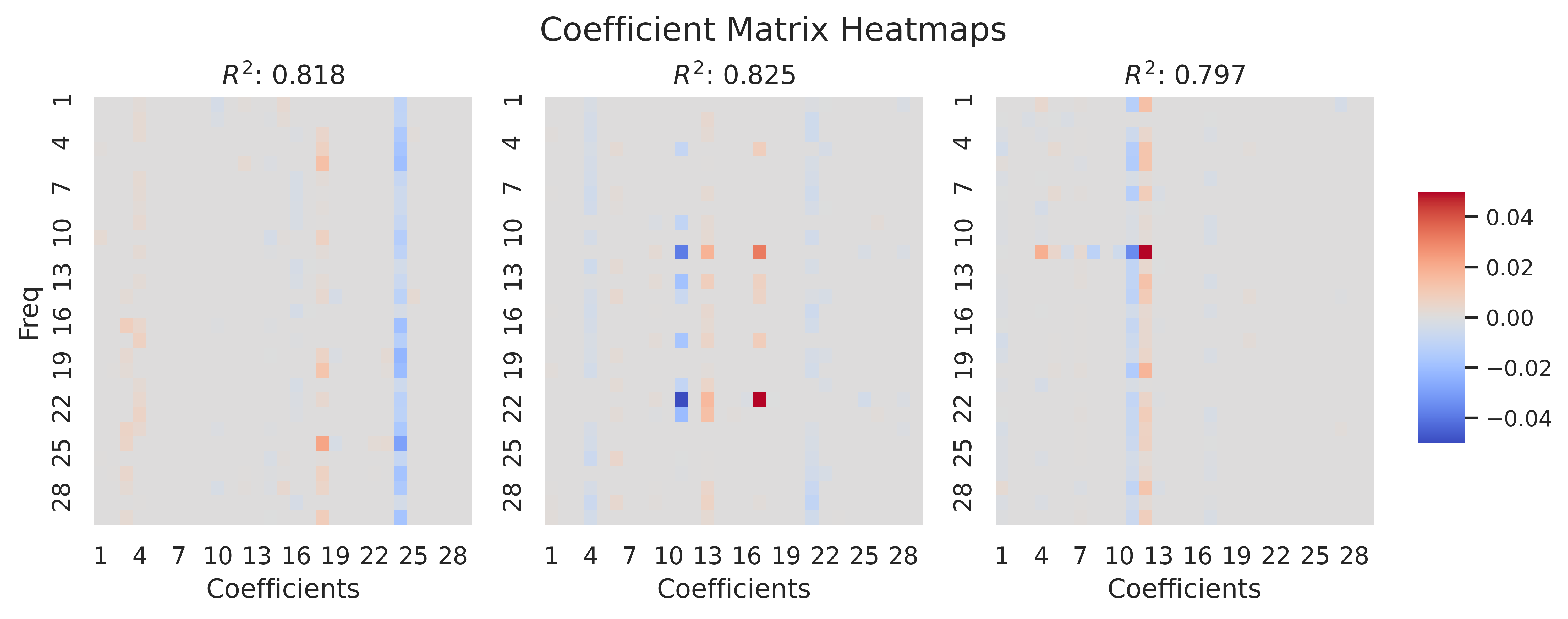}
    \caption{Coefficients heatmaps for 3 more Lasso fits of the embeddings' training dynamics.}
    \label{fig:ode_more_coefs}
\end{figure}

\begin{figure}[htb]
    \centering
    \includegraphics[width=\linewidth]{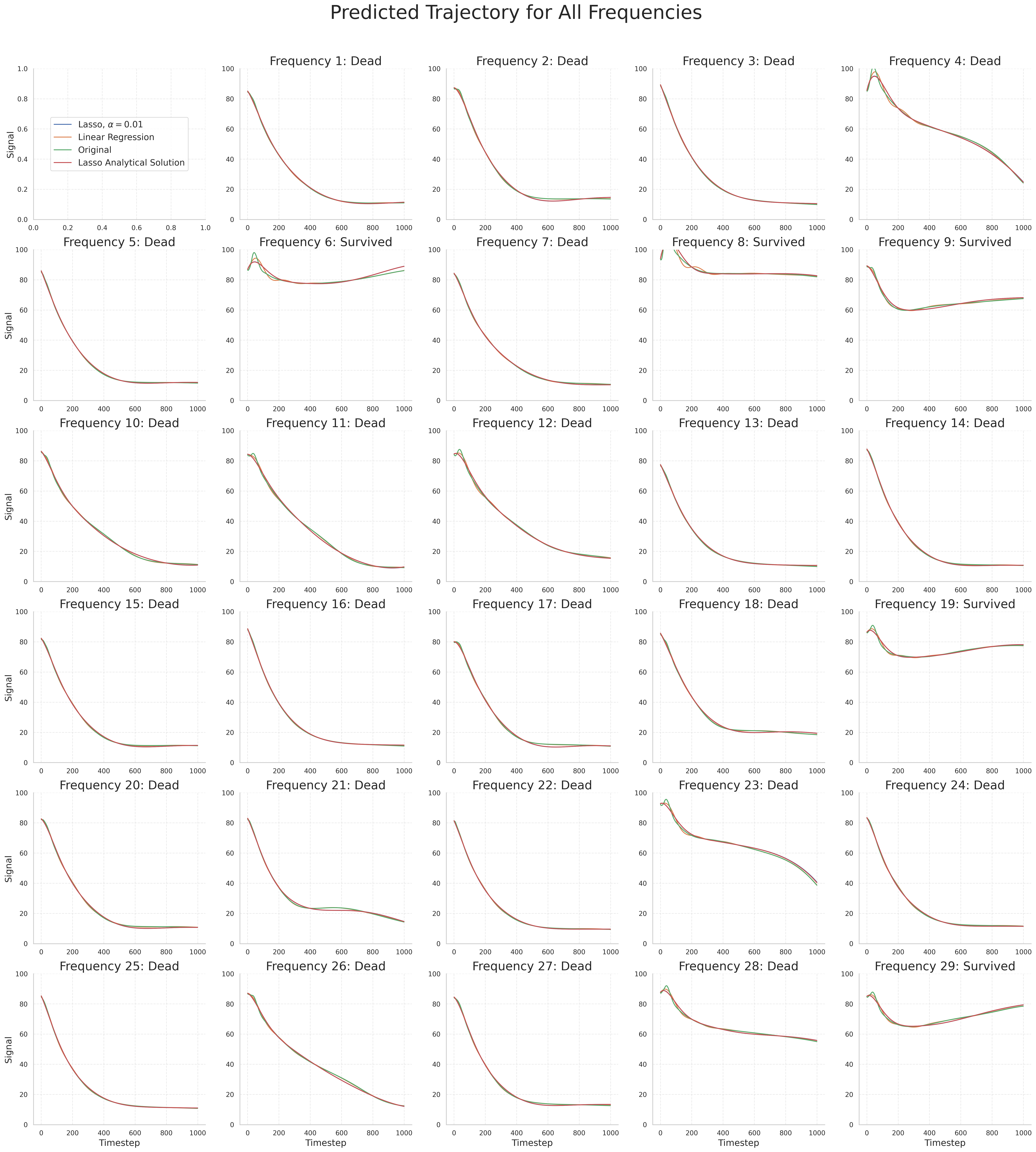}
    \caption{Original and estimated trajectories for signal evolution of all $29$ frequencies during training.}
    \label{fig:ode_all}
\end{figure}

In Figure \ref{fig:ode}, we have shown the trajectories of two representative frequencies approximated by our linear ODE, one dead and one survived, and one representative coefficients heatmap. 
In Figure \ref{fig:ode_more_coefs}, we show the sparse coefficients heatmap for 3 more ODE fits with Lasso. In Figure \ref{fig:ode_all}, we report predicted trajectories for all $29$ frequencies in one training run. 

\section{Experiments Compute Resource}
\label{appendix:compute-resource}

All the model training is performed on NVIDIA V100 GPUs. We provide the GPU specs as follows:
\begin{itemize}
    \item Processor: Intel Xeon Gold 6248
    \item Nodes: 224
    \item Clock Rate: 2.5GHz
    \item CPU cores: 40
    \item Node RAM: 384GB
    \item RAM per core: 9GB
    \item Accelerator type: Nvidia Volta V100
    \item Accelerators(per Node): 2
    \item Accelerator RAM: 32GB
\end{itemize}

The GPU days needed for each experiment are:
\begin{itemize}
    \item Initial gradient/gradient experiment in \ref{sec:which-circles}: 1 GPU day
    \item Freeze embedding experiment in Section \ref{sec:freeze-embed}: 0.8 GPU days
    \item Varying modulus $p$ experiment in Section \ref{sec:trainable-embed}: 15 GPU days
    \item Varying dimension experiment in Section \ref{sec:trainable-embed}: 9 GPU days
    \item Initial signal perturbation experiment in Section \ref{sec:init-signal}: 15 GPU days
    \item Manual construction experiment in Section \ref{sec:init-signal}: 0.5 GPU days
    \item Varying weight experiment in Appendix \ref{appendix:weight_decay}: 2 GPU days
    \item Other small-scale experiments: 1 GPU day
\end{itemize}
Overall, the experiments compute resource adds up to about 45 GPU days. The full research project does not require more compute than the experiments reported in the paper.

\section{Acknowledgement for Online Assets Used}
\label{appendix:existing-assets}
We have utilized several online assets in creating Figure \ref{fig:figure1}. The original illustration of the Fourier Transform can be accessed \href{https://medium.com/the-modern-scientist/the-fourier-transform-and-its-application-in-machine-learning-edecfac4133c}{here}. The icons are sourced from \href{https://thenounproject.com/icon/gaussian-distribution-1425410/}{The Noun Project} and \href{https://www.clker.com/clipart-teacher-10.html}{Clker}.